\documentclass[10pt,twocolumn,letterpaper]{article} 

\usepackage{cvpr}
\usepackage{times}
\usepackage{epsfig}
\usepackage[utf8]{inputenc}
\usepackage{amsmath, amssymb}
\usepackage[linesnumbered,ruled]{algorithm2e}
\usepackage{cite}
\usepackage{graphicx}
\usepackage{float}
\usepackage{tikz}
\usepackage[hang,flushmargin]{footmisc}

\usepackage[pagebackref=true,breaklinks=true,letterpaper=true,colorlinks,bookmarks=false]{hyperref}

\def\cvprPaperID{****} 

\cvprfinalcopy

\DeclareMathOperator{\sign}{sign}

\newcommand{\norm}[1]{\left\lVert#1\right\rVert}

\newcommand{\smallfigure}[2]{\begin{figure}[h]
		\includegraphics[width=\linewidth]{figures/#1.png}
		\caption{#2}
		\label{fig:#1}
\end{figure}}

\newcommand{\largefigure}[2]{\begin{figure*}[h]
		\centering
		\includegraphics[width=\textwidth]{figures/#1.png}
		\caption{#2}
		\label{fig:#1}
\end{figure*}}

\newcommand{\customfigure}[3]{\begin{figure}[h]
		\centering
		\includegraphics[width=#3]{figures/#1.png}
		\caption{#2}
		\label{fig:#1}
\end{figure}}

\newcommand\blfootnote[1]{%
	\begingroup
	\renewcommand\thefootnote{}\footnote{#1}%
	\addtocounter{footnote}{-1}%
	\endgroup
}

\title{Classifiers Based on Deep Sparse Coding Architectures are Robust to Deep Learning Transferable Examples}

\author{Jacob M. Springer\textsuperscript{1,2}, Charles S. Strauss\textsuperscript{3}, Austin M. Thresher\textsuperscript{3}, Edward Kim\textsuperscript{4}, \and Garrett T. Kenyon\textsuperscript{1}}

\date{November 16, 2018}

\begin{document}

\maketitle
\thispagestyle{empty}

\blfootnote{\textsuperscript{1}Los Alamos National Laboratory, \textsuperscript{2}Swarthmore College, \textsuperscript{3}New Mexico Consortium, \textsuperscript{4}Villanova University}

\begin{abstract}
Although deep learning has shown great success in recent years, researchers have discovered a critical flaw where small, imperceptible changes in the input to the system can drastically change the output classification. These attacks are exploitable in nearly all of the existing deep learning classification frameworks. However, the susceptibility of deep sparse coding models to adversarial examples has not been examined.  Here, we show that classifiers based on a deep sparse coding model whose classification accuracy is competitive with a variety of deep neural network models are robust to adversarial examples that effectively fool those same deep learning models.  We demonstrate both quantitatively and qualitatively that the robustness of deep sparse coding models to adversarial examples arises from two key properties.  First, because deep sparse coding models learn general features corresponding to generators of the dataset as a whole, rather than highly discriminative features for distinguishing specific classes, the resulting classifiers are less dependent on idiosyncratic features that might be more easily exploited.  Second, because deep sparse coding models utilize fixed point attractor dynamics with top-down feedback, it is more difficult to find small changes to the input that drive the resulting representations out of the correct attractor basin.
\end{abstract}

\section{Introduction}

With their strong and steadily improving performance on image classification tasks, deep feed-forward convolutional networks (DCNs) are now ubiquitous in virtually all computer vision and machine learning challenges. However, overwhelming evidence shows that DCNs have at least one major flaw: they are susceptible to adversarial attacks. Researchers are able to make specially crafted ``adversarial perturbations'' that when added to a correctly-classified image, produce an image that appears semantically indistinguishable from the original yet are confidently misclassified by DCN models \cite{szegedy1, goodfellow1}. In many cases, these adversarial perturbations can be added to any image within a class to cause the DCN to misclassify the image. Even more alarming, adversarial perturbations are commonly transferable across DCN model architectures \cite{szegedy1, goodfellow1, moosavi1}. In an age in which autonomous cars are ever closer to entering the consumer market, in which deep learning is common in medical applications, and in which facial recognition is used for security purposes, the flaw posed by adversarial examples can have devastating effects if used maliciously. Moreover, attempts to mitigate the problem of adversarial examples have largely failed or render established models significantly less effective \cite{jetley1, tsipras1}. In fact, in practice, increasing robustness to adversarial examples in DCN models necessarily reduces classification performance \cite{jetley1, tsipras1}.

In our work, we sought to develop a framework that would be immune to adversarial attack while maintaining a high level of classification performance. In particular, our goal was to mitigate the risk of a neural network to the transfer of adversarial example attacks and low-pass filtering attacks that have been shown to be effective adversaries against deep feedforward networks \cite{szegedy1, goodfellow1, jetley1}. We believe that demonstrating robustness is an important step forward in computer vision and machine learning research and illustrate a thorough process of testing and experimentation in the body of this paper. As a result of our exploration and analysis, we found that every model we tested was susceptible to transferable adversarial examples, except one. This model, deep sparse coding (DSC) \cite{kim1}, takes a novel, biologically-inspired approach to machine learning, and was immune to these transferable adversarial examples generated to attack other deep learning architectures. Furthermore, we found, and illustrate in our results, that the DSC model’s representation of input is invariant to small perturbations. This robustness stems from the model’s tendency to observe semantically meaningful features of an image, in contrast with the high frequency features that have little or no semantic meaning to humans used by the conventional DCN classifiers.

\section{Background}

\subsection{Single-layer sparse coding}

 To introduce the concept of deep sparse coding, we first consider a single-layer of sparse coding. Intuitively, a single-layer of sparse coding requires learning a set of relatively low-dimensional, nearly orthogonal bases, with the set of all such bases forming an overcomplete \textit{dictionary} that has been optimized for parsimoniously encoding any input drawn from the same distribution on which the model has been trained. The fully trained system acts as an autoencoder that is able to infer a sparse representation of any given input $\mathbf{x}$ which can in turn be used to reconstruct $\mathbf{x}$. From a learned dictionary, the sparse coding algorithm attempts to infer the optimal basis in which to project the input by maximizing reconstruction fidelity while using as few basis vectors as possible.

More formally, we can pose the sparse coding problem as follows: given a set of inputs from some input distribution $\mathbf{x}_1, \ldots, \mathbf{x}_n \sim X$ with associated sparse representations $\mathbf{a}_1, \ldots, \mathbf{a}_n$, which we call \textit{activation vectors}, sparse coding attempts to minimize the average mean squared error of the reconstructions of $\mathbf{x}_i$ using $\mathbf{a}_i$ and a dictionary $\Phi$. Additionally, the $\mathbf{a}_i$ must be sparse. Mathematically, inferring a sparse representations is a minimization problem: \begin{equation} \min_\Phi \sum_{i=1}^n \min_{\mathbf{a}_i} \frac{1}{2} \norm { \mathbf{x}_i - \Phi \mathbf{a}_i }^2_2 + \lambda \norm{\mathbf{a}_i}_0 \end{equation} where $\lambda$ is the sparsity penalty parameter. Note that $\Phi \mathbf{a}_i$ represents the reconstruction of $\mathbf{x}_i$ and $\norm{\cdot}_k$ represents the $\ell_k$ norm.

The way in which we solve for an appropriate activation vector for each $\mathbf{x}$ is inspired by neural biological processes. We represent the internal state of the model with a \textit{membrane potential vector} $\mathbf{u}$ associated with the activations. We can compute $\mathbf{a}$ by a simple thresholding function: \begin{equation} a_m = \begin{cases} 
u_m - c & u_m \geq c \\
0 & \text{otherwise.} 
\end{cases} \end{equation} where $c$ is the thresholding constant.

To minimize reconstruction error while maintaining a desired sparsity, we use a technique known as the locally competitive algorithm \cite{rozell1}, allowing our sparse representation to evolve over time as a dynamical system. We wish to minimize the energy $J$ of our system, which we define as the mean squared error of the reconstruction and a sparsity cost penalty $\norm{\mathbf{a}}_1$. Formally: 
\begin{equation} J = \frac{1}{2} \norm{\mathbf{x} - \Phi \mathbf{a}}^2_2 + \lambda  \norm{\mathbf{a}}_1. \end{equation}
Note that we use an $\ell_1$ norm as an approximation for the $\ell_0$ norm of the activation vector to make the system differentiable. We can express the time evolution rule as a differential equation obtained by taking the negative gradient of the energy $J$ with respect to $\mathbf{a}$, \begin{equation} \dot{\mathbf{u}} = -\mathbf{u} + \Phi^T \mathbf{x} - [\Phi^T \Phi \mathbf{a} - \mathbf{a}] \end{equation} which can be described as the combination of three intuitive components. The $-\mathbf{u}$ term acts as the decay component of the system, thus slowly decreasing the potential of each neuron over time. If an active neuron is not continuously excited, then it will quickly fall below threshold and deactivate. The $\Phi^T \mathbf{x}$ term ``charges up'' each of the neurons, exciting neurons whose features effectively match the input. The $[\Phi^T \Phi \mathbf{a} - \mathbf{a}]$ term acts as an inhibitory signal, causing neurons that explain a similar component of the data to compete, thus allowing only the neurons that ``best'' explain the data to remain active. Rozell \etal \cite{rozell1} proves that this dynamical system converges to the optimal solution to the energy function.

\subsection{Multi-layer (deep) sparse coding}

Analogous to deep feed-forward models, deep sparse coding simply adds more hidden layers between the input and output layers, where each hidden layer attempts to reconstruct the layer immediately below it. However, in contrast to conventional deep learning models, the data connections are not simply unidirectional, forward connections from one layer to the next. Instead, deep sparse coding integrates multiple highly recurrent connections at every level, i.e., bottom-up input, lateral, and top-down feedback signals. The top-down feedback connections reinforce a stable and self-consistent representation of the data at each layer. 

To illustrate the necessity of top-down feedback, consider that a typical feed-forward network might activate multiple ``good'' explanations of the data. For example, the neurons in the early layers of a model only perceive a small patch of the input due to their local and limited receptive fields. These neurons can easily observe cat-like features and dog-like on the same input. A feed-forward model might encode this input by activating both neurons that respond to cat-like features and those that respond to dog-like features. This is because it is nearly impossible for bottom-up and lateral processes to ``know'' which low-level feature should be assigned to one shape vs. another. Thus, feedback projections from higher-level regions would be necessary to account for our visual abilities in all but the simplest of circumstances \cite{kveraga1}. On the other hand, consider the case when a higher-level neuron that activates on images of dogs becomes active in the DSC model. If the model had learned that observing a dog reduces the probability of observing a cat, then the top-down feedback from the higher-level dog neuron will inhibit the neurons that respond to cat-like features and would reinforce the neurons that respond to dog-like features.

Thus, top-down feedback provides an effective method to converge on a confident and self-consistent representation of the input. The DSC model formulates top-down feedback as follows: Let $\mathbf{u}_\mathcal{L}$ represent the internal state of layer $\mathcal{L}$. Let the residual $\mathbf{r}_{\mathcal{L}+1}$ of layer $\mathcal{L}+1$'s reconstruction of layer $\mathcal{L}$ be defined: \begin{equation} \mathbf{r}_\mathcal{L} = \mathbf{u}_\mathcal{L} - \Phi_{\mathcal{L}+1} \mathbf{a}_{\mathcal{L}+1} \end{equation}

The $\mathbf{u}_\mathcal{L}$ term is forced to ``conform'' to layer $\mathcal{L}+1$'s reconstruction of it by modifying our time evolution function of $\mathbf{u}_\mathcal{L}$ to continually reduce the residual of the higher-level reconstruction: \begin{equation} \dot{\mathbf{u}}_\mathcal{L} = -\mathbf{u}_\mathcal{L} + \Phi^T \mathbf{x} - [\Phi^T \Phi \mathbf{a}_\mathcal{L} - \mathbf{a}_\mathcal{L}] - \mathbf{r}_\mathcal{L} \end{equation}

\section{Methodology}

In this paper, we begin by defining a binary classification task that will be attacked using adversarial examples. We chose the binary classification task of face identification as this is an important domain where a machine learning model should be robust. The face ID application will be applied to a set of deep feed-forward convolutional networks as well as on the deep sparse coding framework. Earlier work on the deep sparse coding framework used the example of ``Halle Berry'' face detection \cite{kim1} and thus we continue on with this theme. We first demonstrate that we can train a number of DCNs that can accurately distinguish between images of Halle Berry from our dataset and images that are not of Halle Berry. Next, we exploit the learned ``Halle Berry neuron'' in the DSC model, i.e., a neuron analogous to the neuron in the human brain identified by \cite{koch1} that responds to the multimodal ``concept'' of Halle Berry. After training the DCN models, we show that we can generate adversarial examples from our holdout set that transfer across DCN architectures using established techniques. Finally, we show that these adversarial examples fail to transfer to the DSC model.

\subsection{Dataset}

\largefigure{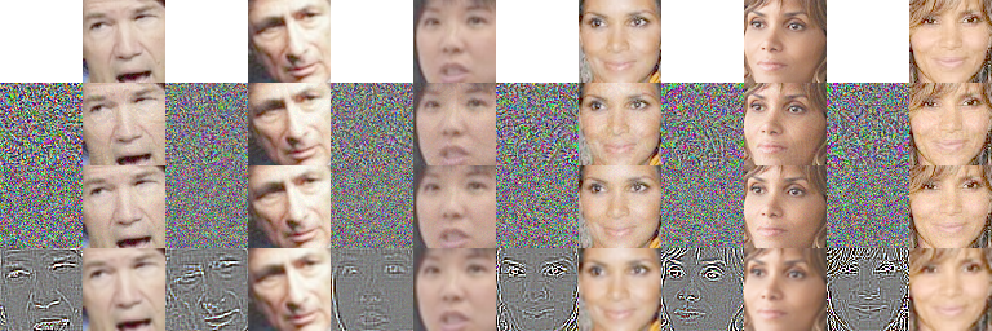}{Images from the holdout dataset and the corresponding alterations. In each row, the left three faces are of ``other'' (non-Halle Berry), and the right three faces are of Halle Berry. \textit{From top to bottom}: original, adversarial, Gaussian noise, low-pass filter. The noise to the left of each non-original image shows the difference between that image and the corresponding original. The noise has been translated so that a pixel with no change is gray with a value $0.5$. The noise is scaled by a factor of $10$ in order to increase visibility. While upon close inspection the alternations are visible, each altered image is nearly indistinguishable identical to the original image.}

We use a subset of the Labeled Faces in the Wild (LFW) dataset \cite{huang2} augmented with images of Halle Berry. The dataset consists of photographs of faces from the web detected by the Viola-Jones face detector. In total, there are 370 images of Halle Berry, scraped from IMDB, and a total of 5,763 images consisting of 2,189 unique people. Each image is $64 \times 64$ pixels, each with $3$ color channels. We segment the data into 4,519 training images, 1094 validation images, and 150 holdout images for evaluating our final models. We use 150 holdout images which includes 75 images of Halle Berry and 75 images of other people. Each of these holdout images are altered to make the variants of the dataset as described in later sections, including the adversarial dataset, the low-pass filtered dataset, and the Gaussian noisy dataset. See Figure \ref{fig:dataset} for a sample of the dataset and its variants. As a note, all figures in this paper are best viewed in color.

\subsection{Deep feed-forward convolutional network models}

Deep convolutional neural networks (DCNs) that have been pre-trained on large image datasets such as ImageNet \cite{russakovsky1} and then fine-tuned on the target dataset have been shown to have impressive classification performance on several datasets of limited size \cite{donahue1, razavian1}.  Thus, we initialized several state-of-the-art DCN architectures (ResNet50 \cite{he1}, InceptionV3 \cite{szegedy2}, VGG16 \cite{simonyan1}, MobileNetV2 \cite{sandler1}, DenseNet121 \cite{huang1}) with weights trained on ImageNet \cite{tensorflow, keras, huang1, he1, szegedy2, simonyan1, sandler1} and then fine-tuned these DCNs to distinguish between images of Halle Berry and images of ``other'' faces. We used the Keras \cite{keras} framework on top of TensorFlow \cite{tensorflow} to build the models.

\subsection{Deep sparse coding model}

\smallfigure{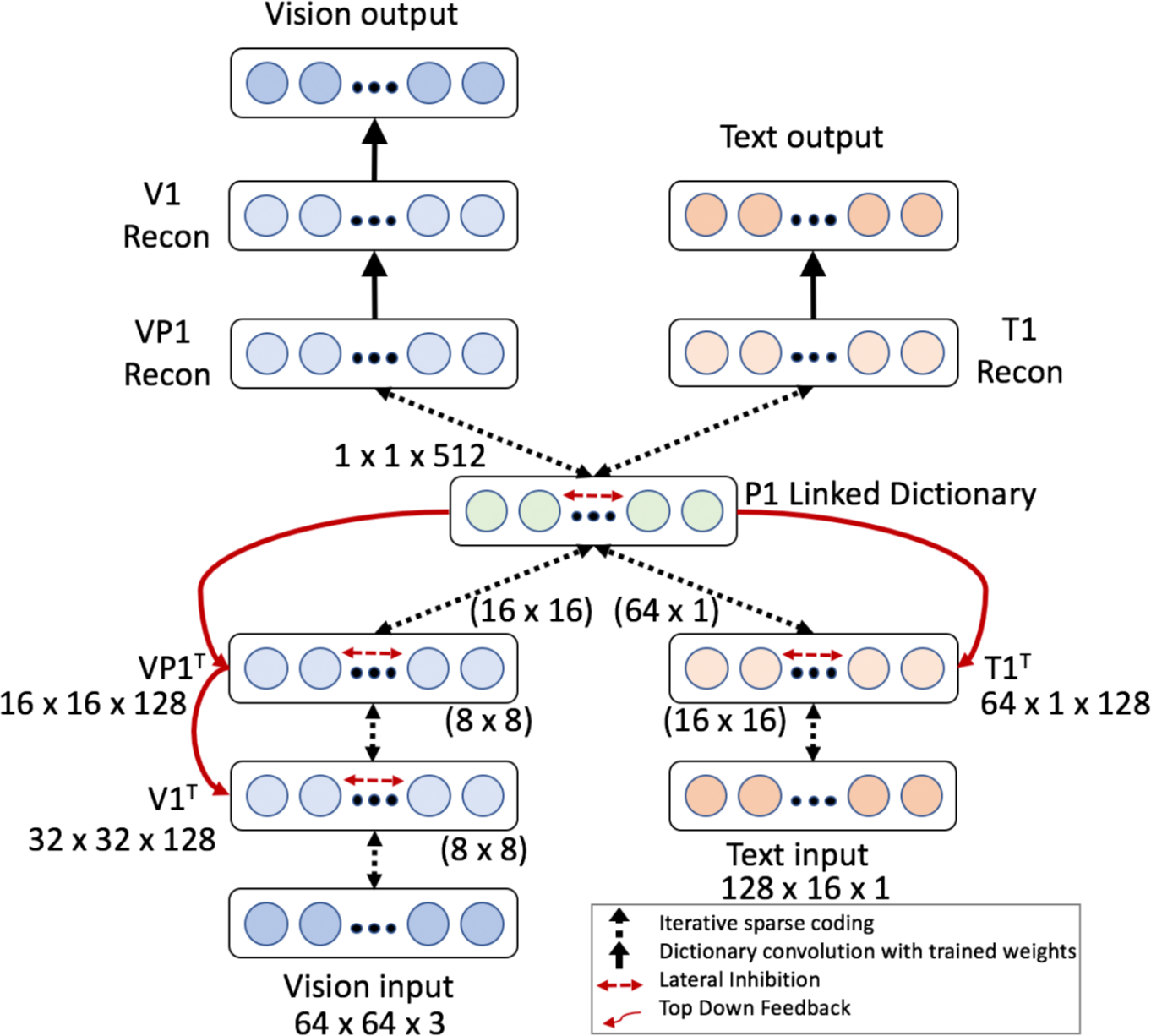}{Deep sparse coding (DSC) model from Kim \etal \cite{kim1}}

We use the DSC model built and trained by Kim \etal described in \cite{kim1}.  The model attempts to reconstruct two modalities using a single shared top-level hidden layer. The two modalities are a $64 \times 64$ color image of a person's face and a $128 \times 16$ grayscale image of the person's name, rendered in 20 point Arial font. The bi-modality of the model with a shared P1 layer (see Figure \ref{fig:model}) encourages the model to encode an image of a person and an image of their name with the same representation. We show that this feature encourages the model to learn high-level features including a ``Halle Berry neuron'', i.e., a single neuron which appears to encode both images of Halle Berry and of her name. The model is implemented using PetaVision \cite{petavision}, an open-source accelerated neuromorphic computing framework.

The full model is depicted in Figure \ref{fig:model}. The model consists of two vision-specific layers, V1 and VP1, and one text-specific layer, T1.  These recurrent layers are connected through the top-level layer, P1. For a more detailed description of the model, see \cite{kim1}.

The DSC model does not perform classification explicitly. Rather, it generates a sparse representation of the input in its P1 layer. For classification, we train a linear support vector machine (SVM) to take input from the P1 layer activations associated with a given image and output a label.

\subsection{Transferable adversarial examples}

We demonstrate that \textit{transferable adversarial examples}, i.e., adversarial examples that generalize across multiple DCN architectures, do not generalize to the DSC model. To generate an adversarial example for input $\mathbf{x}$ on classifier $\hat{k}$, we wish to find perturbation $\mathbf \eta$ such that $\hat k(\mathbf{x}) \neq \hat k(\mathbf{x} + \mathbf{\eta})$ subject to the constraint that $\norm{\eta}_\infty$ is sufficiently small. We generate adversarial examples using an iterative version of the fast gradient sign method \cite{goodfellow1} (IFGSM). Let $J(\mathbf{x}, y)$ be the classification loss function given input $\mathbf{x}$ and ground-truth label $y$. Beginning from the original input $\mathbf{x}^*_{0} = \mathbf{x}$, we describe the algorithm as follows: \begin{equation} \mathbf{x}^*_{i+1}=\mathbf{x}^*_{i} + \varepsilon \, {\sign \nabla_\mathbf{x} J(\mathbf{x}^*_{i}, y)} \end{equation} 

Low-pass filtering has been shown to reduce classification performance on DCN architectures \cite{jetley1}. To demonstrate that the DSC model does not rely on high-frequency noise to classify images, we run each holdout image through a low-pass filter to generate a fourth dataset. 

Samples of these datasets are shown in Figure \ref{fig:dataset}.

\section{Experiments and results}

We first illustrate the responses of the Halle Berry neuron.  To demonstrate the DSC model's resistance to transferable adversarial examples, we evaluate the DSC model against analogous DCN models by comparing Halle-Berry-identification performance on original images, adversarial images, Gaussian noisy images, and low-pass filtered versions.  We examine the internal state of the DSC model to identify the properties of the DSC model that provide its robustness.

\subsection{Representation of Halle Berry}

\customfigure{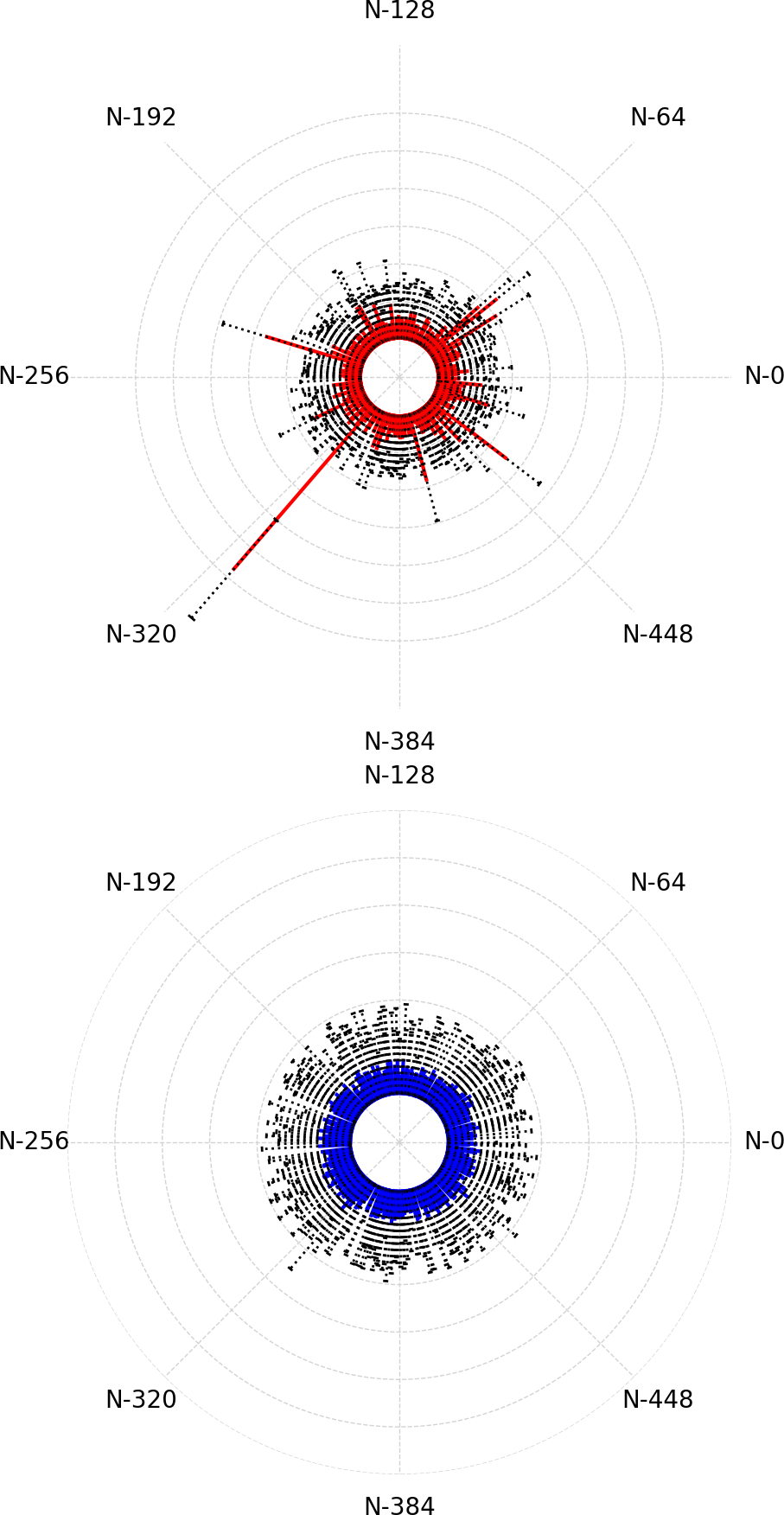}{Average activations of P1 layer over the training set. Standard deviations are plotted with a dotted black line. \textit{In red}: average response over all images of Halle Berry. \textit{In blue}: average response over all other faces. On average, the P1 layer has a strong spike on \textit{N-326} for Halle Berry images and comparatively weak activations to tther faces. The average activations over all other faces is relatively uniform. Both graphs are plotted on the same scale.}{200px}

\smallfigure{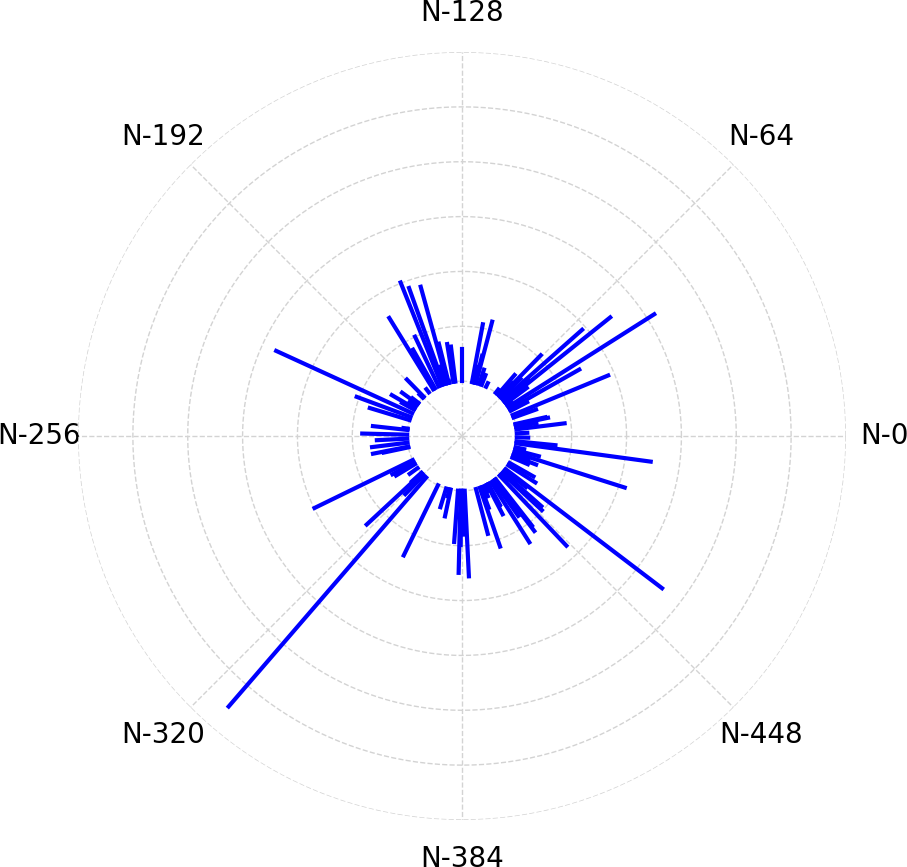}{Positive weights of an SVM trained to classify images as either ``Halle Berry'' or ``other'' from the sparse P1 layer representation of each image. Importantly, the strongest positive weight corresponds to \textit{N-326}, indicating that this neuron acts as the strongest linear indicator of Halle Berry. Since none of the negative weights have a magnitude close to that of the weight that corresponds to \textit{N-326}, we omit them to improve the clarity of the diagram.}

We observe a single ``Halle Berry neuron'', which we call \textit{N-326} due to its index in the DSC model. To demonstrate the selectivity of \textit{N-326}, we compare the average activation of all P1 neurons over all images in the training set, and over images of Halle Berry in the training set. We plot these radially in Figure \ref{fig:activationserrorbars}. The average activation of \textit{N-326} on images of Halle Berry is significantly stronger than the average activation produced by other faces, suggesting that this neuron activates strongly and selectively in response to an image of Halle Berry. In fact, thresholding \textit{N-326} yields a classification performance of nearly 85\% on the holdout dataset. Thus, \textit{N-326} acts as a somewhat effective indicator of images of Halle Berry. Additionally, after training an SVM on the full P1 layer to distinguish images of Halle Berry from other faces, we examine the weights, which are plotted in Figure \ref{fig:svmweights}. The weight associated with \textit{N-326} is nearly twice the second-highest weight, further suggesting that \textit{N-326} is the strongest linear indicator of Halle Berry. 

\subsection{Comparison of models}

\smallfigure{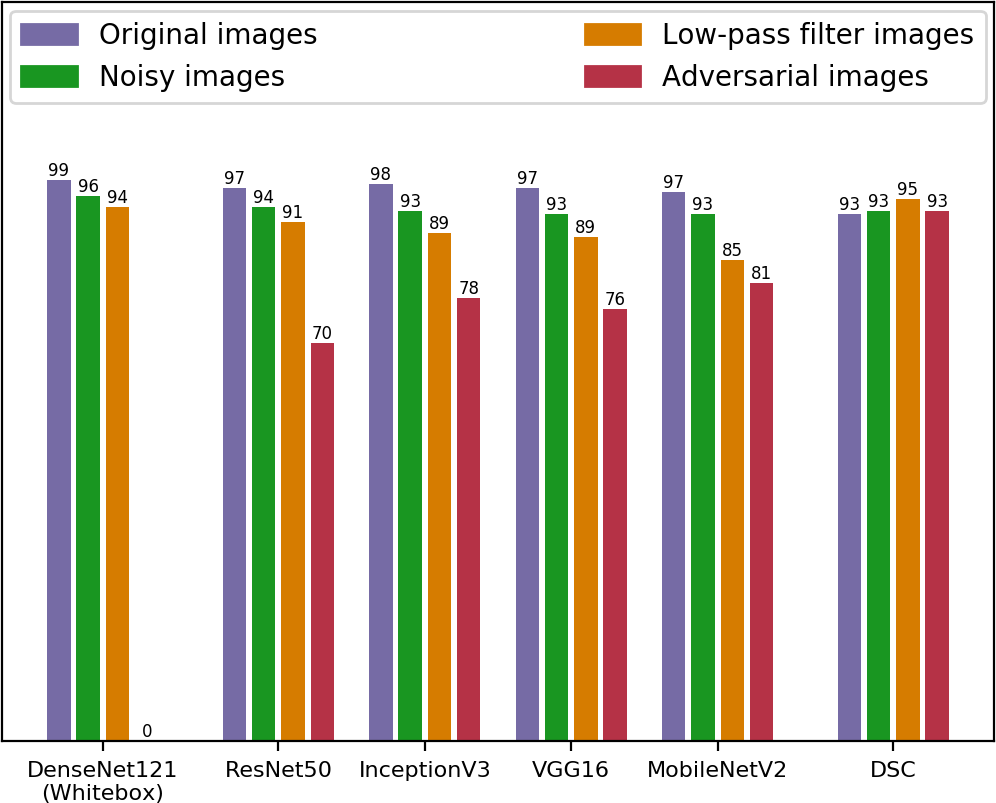}{Accuracy scores of the various models on the \textit{in blue:} original images, \textit{in green:} Gaussian-noisy images, \textit{in orange:} low-pass filtered images, and \textit{in red:} adversarial images. Note that we use the DenseNet121 model to generate the adversarial examples, which is why its classification accuracy on the adversarial examples is zero. We see that the performance of all DCN models  drops significantly on adversarial examples whereas the performance of the DSC model is invariant to the described minor alterations.}


We compare robustness  by measuring the performance of each model on the holdout datasets. We entertain the possibility that any distortion of an image might cause a model to be unable to identify any particular features that correlate strongly with images of Halle Berry or of other faces, which would force the network to guess randomly. Thus, we control for the effects of the distortion by running the model on the input with added Gaussian noise with the same mean and standard distribution as the adversarial noise. The comparative accuracies are plotted in Figure \ref{fig:accuracies}. Note that the performance of the DenseNet121 model on the adversarial examples is shown for reference, however, since the DenseNet121 model is used as a whitebox with which to generate adversarial examples, the performance on adversarial examples is expected to be near-zero and orders of magnitude smaller than the performance of the other DCN architectures on the transferred adversarial examples, corresponding to a blackbox attack.

The classification performances of the respective models demonstrate the robustness of the DSC model to transferable adversarial examples. The four applicable DCN models (ResNet50, InceptionV3, VGG16, and MobileNetV2) perform significantly worse on the transferred adversarial examples than on the original and Gaussian noisy images, yet the difference in the performance of the DSC model is negligible between the original and adversarial inputs. 

Furthermore, the classification performance of the DCN models decreased on images after processing them with a low-pass filter. However, as shown by Figure \ref{fig:accuracies} the performance of the DSC model is invariant to a low-pass filter.

\subsection{Invariance of representation}

\begin{table}[tbh]
	\small
	\begin{tabular}{llllll}
		&ResNet & Inception & VGG & MobileNet & \textbf{DSC} \\
		$\delta_{\alpha}$&0.44   &    0.08       &    0.59     &    0.68      &     \textbf{0.05}        \\
		$\delta_{\eta}$&0.34   &    0.08       &    0.42    &     0.69      &     \textbf{0.04}      \\
		$\delta_{\ell}$&0.39   &    0.13        &    0.44   &     0.46      &      \textbf{0.04}  
	\end{tabular}
	\caption{Distance metric between hidden-layer representation of original images and adversarial images ($\delta_\alpha$), Gaussian noisy images ($\delta_\eta$), and low-pass filter images ($\delta_\ell$). The exact metric is described below. Deep sparse coding representations of adversarial, noisy, and low-pass filter images are significantly more similar than the corresponding representations by DCN models.}
	\label{tab:representationcomparisons}
\end{table}

\largefigure{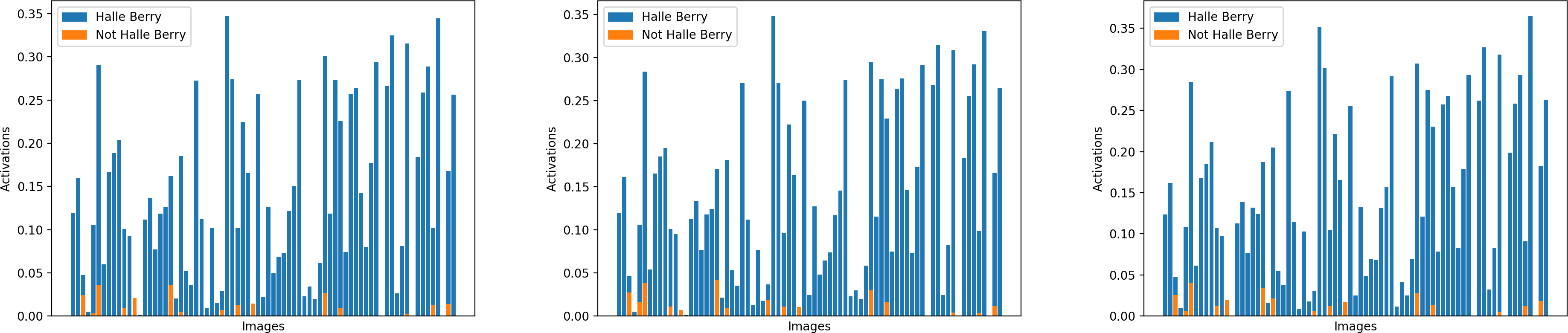}{The activation strength of the \textit{N-326} neuron on each holdout image, compared across the variants of the datasets. The left graph shows the \textit{N-326} activation on each \textit{in blue:} original image of Halle Berry and \textit{in orange:} of people other than Halle Berry. The center and right graphs show the same, but on the adversarial copies and on the low-pass filter copies, respectively. \textit{N-326} activates strongly on images of Halle Berry and weakly on images of other. The \textit{N-326} activation remains invariant across adversarial perturbations and low-pass filtering.}

\smallfigure{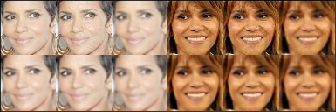}{Representative examples of reconstructions of Halle Berry faces and corresponding alterations. The images provided as input to the model are shown on the top row. The bottom row features the model's reconstruction of each image. For each face, from left to right: original, adversarial, and low-pass filter.}

\customfigure{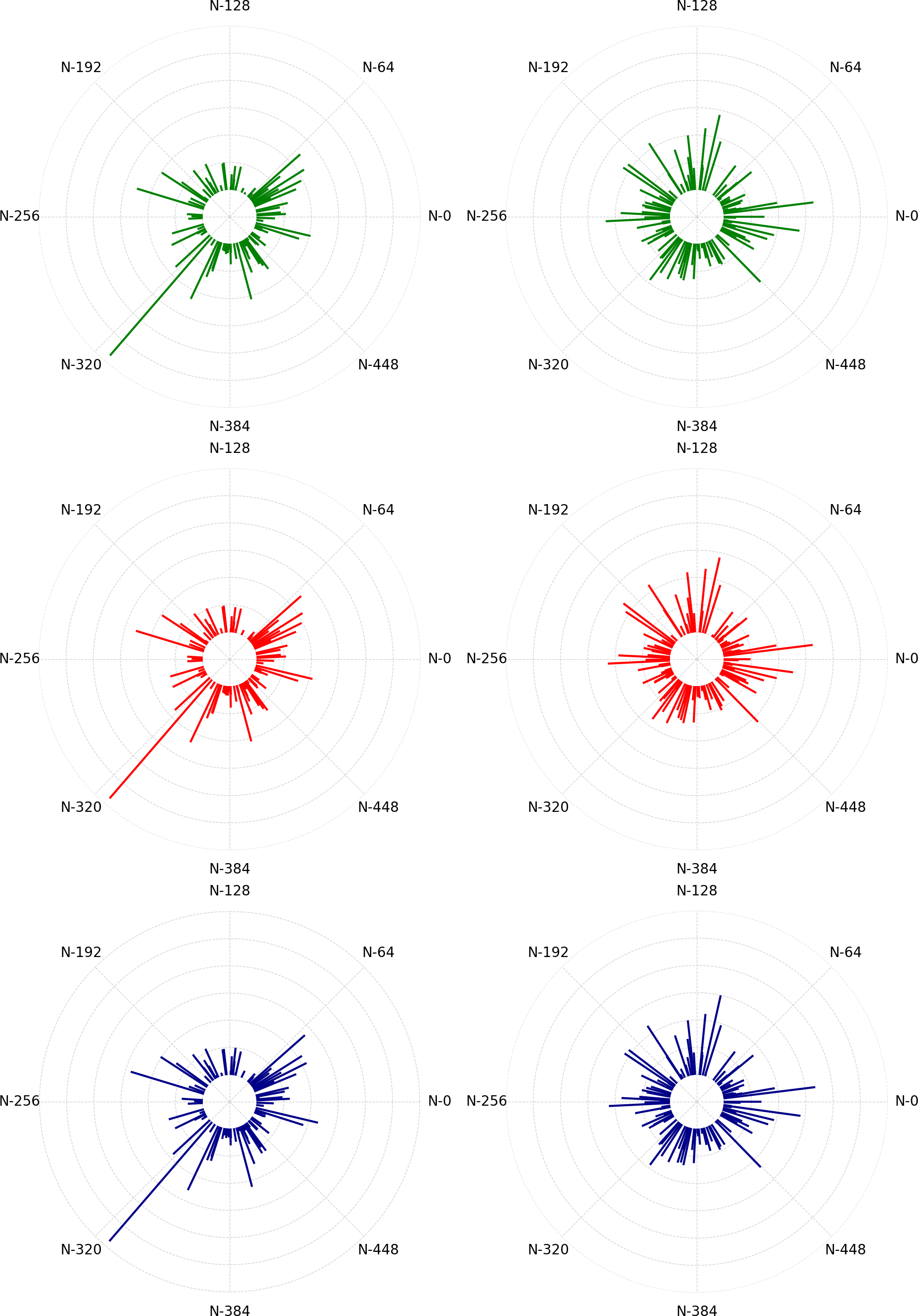}{P1 layer activations on Halle Berry and not Halle Berry across perturbations. The \textit{left column} shows the activations of all 512 neurons in the P1 layer of the DSC model on an arbitrarily selected image of Halle Berry from our holdout set. The \textit{right column} shows the activations of the P1 layer on an arbitrarily selected image of a person's face who is not Halle Berry. The \textit{top row} shows activations on the original version of the image; the \textit{center row} shows activations on the adversarial version of the image; the \textit{bottom row} shows the activations on the original image after being processed by a low-pass filter.}{230px}

To show that the DSC model is resistant to minor high-frequency perturbations, we extract and plot the activation of the strongest linear indicator of Halle Berry, \textit{N-326}, on the 75 holdout images of Halle Berry and the 75 holdout images of other. We provide three graphs, placed adjacently, that show the \textit{N-326} activation on the original images, the adversarial images, and the low-pass filter images, respectively. These graphs are shown in Figure \ref{fig:n326activations}. Adversarial perturbations and low-pass filtering have little effect on the activation of \textit{N-326}. It follows that any linear classifier that relies heavily on \textit{N-326}, such as the SVM we use to classify the images, should be resistant to these minor perturbations, including alterations that prove adversarial in feedforward DCNs.

We further examine the P1 representation of images across the original images, adversarial perturbations, and a low-pass filter. In Figure \ref{fig:activations}, we plot the P1 activations of a representative image of Halle Berry (top) and of other (bottom). The representation of each image is nearly identical across the alterations. In other words, these minor perturbations do not appear to affect the entire P1 representation of an image.

These observations generalize to the entire holdout dataset. We compare the top-level hidden-layer representation of each original input image with the corresponding representation of the adversarial, Gaussian noisy, and low-pass filter images. As a measure, we compute the similarity between the neuron activations in the hidden layer immediately below the output layer in all networks. In the DSC model, we observe the P1 layer activations when computing the similarity metric. To control for the various distributions of neuron activations across networks when measuring similarity, we compute the average difference between the neuron activations over the average standard deviation of the neuron activations. Shown in Table \ref{tab:representationcomparisons}, we demonstrate that the DSC model has a significantly more similar representations across the original, adversarial, and low-pass filter images than the DCN models we trained. Thus, minor perturbations to the input do not affect the internal representation of the image in our P1 layer.

For completeness, we include representative examples of the reconstructions of the inputs from the three datasets: original, adversarial, and low-pass filter. As illustrated by Figure \ref{fig:reconstructions}, there is little, if any, noticeable difference across the reconstructions of the images in each dataset.

\subsection{Limitations}

We believe that deep sparse coding makes an important step forward towards the development of machine learning models that are fully robust to adversarial examples. However, as with any novel approach, deep sparse coding has not yet been tested on many datasets. 

While we show that transferable adversarial examples generated using the iterative fast gradient sign method do not transfer to the DSC model, we do acknowledge that the DSC model is susceptible to adversarial attacks. This is no surprise since all visual recognition systems (including biological ones) will eventually change their classification given a strong enough perturbation to the input signal. Since observing the gradient of the DSC model is non-trivial, we do not perform a direct gradient-based attack on the DSC model, despite the possibility that an effective one might exist. However, we are able to estimate the gradient of the model by making single-pixel changes to a given image and aggregating them to construct a first-order estimation of the gradient. These examples prove effective against the DSC model and their properties should be examined in the future. This paper, however, still presents an important result: the DSC model is not susceptible to transferable adversarial examples that effectively target DCN models.

\section{Conclusion}

We show that the deep sparse coding model is resistant to transferable adversarial attacks that harshly drop performance of state-of-the-art DCN models. To demonstrate this, we build state-of-the-art DCN classifiers that apply transfer learning to separate images into the categories of Halle Berry and other. We apply the iterative fast gradient sign method to the DenseNet model to generate adversarial examples, and then show that these adversarial examples transfer more effectively than random-noise perturbations to other DCN models. We then show that the DSC model achieves the same classification rate on adversarial examples and low-pass filter images as the original unperturbed images. Additionally, we show that the internal DSC representation of each image is invariant to minor perturbations. Finally, we acknowledge that adversarial examples exist for the DSC model, but that traditional transfer attacks fail on the DSC model.

At a high level, we demonstrate that the deep sparse coding model relies on features that are \textit{different} than those relied upon by DCN models, which has important implications for the field of adversarial machine learning. Despite the fact that the DSC model is vulnerable to direct adversarial attacks, if an attacker does not have access to the specific model, an attack may be difficult to perform. For example, an adversary attempting to fool a self-driving car without access to the model would not be able to apply a standard adversarial example transfer attack. A malicious doctor could not apply a transfer attack to fool the model into identifying a disease from an X-ray that is not present.

We show that the DSC model is significantly more resistant to high-frequency alterations of the data (such as Gaussian noise and DCN adversarial examples) than traditional DCN models. As such, the model may be especially effective in environments in which high-frequency noise occurs frequently. Furthermore, we show that a low-pass filter does not affect the DSC model's internal representation of the data, suggesting that the DSC model relies on lower-frequency features than DCN models.

In our future work, we plan to further investigate deep sparse coding models, especially to answer questions such as: what particular attacks are effective against DSC models, do there exist universal adversarial examples as found in DCN models \cite{moosavi1}, and do adversarial examples effectively transfer across separate DSC models?

In conclusion, we believe that the DSC model provides an important step forward in the field of computer vision and machine learning, and warrants further research. Our project addresses a major flaw in the traditional deep learning methodologies and offers an effective alternative. Our hope is to see further research on neurally inspired models and their applications in the pattern recognition community.

\bibliographystyle{unsrt}
\bibliography{references}

\end{document}